\documentclass[conference,letterpaper]{IEEEtran}

\IEEEoverridecommandlockouts

\usepackage{cite}
\usepackage{amsmath,amssymb,amsfonts}

\usepackage{algorithm}
\usepackage{algpseudocode}

\usepackage{graphicx}
\usepackage{textcomp}
\usepackage{xcolor}

\usepackage{float}
\usepackage{url} 

\usepackage{vcell}
\usepackage{balance} 

\usepackage[utf8]{inputenc}
\usepackage[T1]{fontenc}
\usepackage{pdflscape}

\usepackage{array}
\usepackage{booktabs}
\usepackage{soul} 
\usepackage{siunitx}
\usepackage{numprint}
\usepackage{tabularx}
\usepackage{xltabular}
\usepackage{makecell}
\usepackage{multirow}
\usepackage{adjustbox}

\usepackage{caption}
\usepackage{subcaption}

\usepackage{mweights}
\usepackage{cleveref}
\usepackage{bchart}
\usepackage{stfloats}

\usepackage{tikz}
\usepackage{pgfplots}
\usepackage{pgfplotstable}
\pgfplotsset{compat=1.18}

\def\BibTeX{{\rm B\kern-.05em{\sc i\kern-.025em b}\kern-.08em
    T\kern-.1667em\lower.7ex\hbox{E}\kern-.125emX}}


\author{\IEEEauthorblockN{Farzana Akter\textsuperscript{1,*},
Rakib Hossain\textsuperscript{2,*},
Deb Kanna Roy Toushi\textsuperscript{1},
Mahmood Menon Khan\textsuperscript{1}, \\
Sultana Amin\textsuperscript{2}, and
Lisan Al Amin\textsuperscript{2,\textdagger}}
\IEEEauthorblockA{\textsuperscript{1}Dept. of Information Technology, Washington University of Science and Technology, 
Alexandria, VA, USA}
\IEEEauthorblockA{\textsuperscript{2}Cognitive Links LLC,  
Maryland, USA}
\IEEEauthorblockA{*\textit{These authors contributed equally to this work}}
\IEEEauthorblockA{\textdagger\textit{Corresponding author: lisan@cognitivelinks.com}}}

\begin{document}

\title{
\vspace*{1mm} Hybrid Federated and Split Learning for Privacy Preserving Clinical Prediction and Treatment Optimization}

\maketitle

\begin{abstract}
Collaborative clinical decision support is often constrained by governance and privacy rules that prevent pooling patient-level records across institutions. We present a hybrid privacy-preserving framework that combines Federated Learning (FL) and Split Learning (SL) to support decision-oriented healthcare modeling without raw-data sharing. The approach keeps feature-extraction trunks on clients while hosting prediction heads on a coordinating server, enabling shared representation learning and exposing an explicit collaboration boundary where privacy controls can be applied. Rather than assuming distributed training is inherently private, we audit leakage empirically using membership inference on cut-layer representations and study lightweight defenses based on activation clipping and additive Gaussian noise. We evaluate across three public clinical datasets under non-IID client partitions using a unified pipeline and assess performance jointly along four deployment-relevant axes: factual predictive utility, uplift-based ranking under capacity constraints, audited privacy leakage, and communication overhead. Results show that hybrid FL-SL variants achieve competitive predictive performance and decision-facing prioritization behavior relative to standalone FL or SL, while providing a tunable privacy-utility trade-off that can reduce audited leakage without requiring raw-data sharing. Overall, the work positions hybrid FL-SL as a practical design space for privacy-preserving healthcare decision support where utility, leakage risk, and deployment cost must be balanced explicitly.
\end{abstract}


\section{Introduction}
Clinical machine learning \cite{_24} is increasingly expected to move beyond proof-of-concept models trained at a single hospital and toward systems that generalize across sites, populations, and care pathways. However, the data needed for broad generalization are rarely co-located. Privacy rules, contractual constraints, and operational realities often make central pooling impractical, while harmonization, legal review, and security approval add time and cost. As a result, collaborative training without moving raw patient data has become a practical requirement.

Federated learning (FL) is a leading paradigm for this setting because it keeps patient records within each organization and exchanges model updates under a coordination protocol \cite{_1}. Reviews in health care emphasize both FL's promise and persistent barriers to deployment, particularly under non-identically distributed clinical populations and heterogeneous infrastructure \cite{_2}. Split learning (SL) offers a complementary trade-off by partitioning a network between client and server so that intermediate representations, rather than raw data, cross the boundary. Hybrid designs that combine FL-style aggregation with SL-style partitioning can reduce client-side compute, mitigate stragglers, and provide governance controls over what information is exposed \cite{_5,_6}. At the same time, clinical decision support often requires estimates of incremental benefit for a patient, not only risk prediction. This need is naturally captured by heterogeneous treatment effect estimation and uplift modeling, which target differences between potential outcomes under treated and untreated conditions \cite{_15}.

In this paper, we propose and evaluate a privacy-preserving collaborative framework that combines FL and SL for joint disease prediction and treatment optimization. Unlike many healthcare FL studies that focus primarily on predictive accuracy, we organize evaluation around three practical requirements: reliable prediction and calibration, uplift-based prioritization under limited treatment capacity, and explicit privacy stress testing via membership inference auditing, reflecting evidence that update exchange and intermediate representations can leak sensitive information \cite{_4}. The framework is modular and is assessed across multiple datasets and client partitions using a consistent pipeline, enabling cross-dataset validation rather than a single-benchmark result.

\subsection*{Contributions}
We develop a hybrid FL--SL protocol for multi-institution clinical learning in which cut-layer placement and federated aggregation jointly determine compute placement, communication cost, and exposure at the collaboration boundary. To our knowledge, this is the first study to evaluate a hybrid FL--SL pipeline that simultaneously targets uplift-based treatment optimization and systematic membership inference auditing, validated across multiple clinical datasets. We further report cross-dataset results with consistent preprocessing, partitions, and metrics, including worst-client behavior under non-IID conditions, and we provide an auditable privacy--utility profile that links security signals, practical defenses, and deployment costs.

\section{Literature Review}

This section situates our work across collaborative clinical training, practical privacy auditing, and decision-focused uplift modeling for treatment prioritization.

\subsection{Federated learning in health care}
Federated learning (FL) enables multi-site clinical modeling, but real deployments must handle heterogeneity, partial participation, and communication limits, and locality alone does not guarantee privacy \cite{_1}. Reviews in healthcare report rapid growth in FL studies, yet fewer deployments that address governance, workflow integration, and rigorous multi-center evaluation \cite{_2}. For structured EHR data, surveys highlight additional failure modes beyond imaging, including coding drift, missingness mechanisms, and site-specific practice patterns \cite{_3}. These findings motivate approaches that treat heterogeneity and privacy as first-order objectives.

Governance also shapes what is feasible in practice. Scoping work emphasizes how federations are organized, who controls model artifacts, and how accountability is maintained across institutions \cite{_13}. For hybrid FL--SL, the partition point directly affects what each party can observe and which operating procedures are acceptable.

\subsection{Split learning and hybrid FL--SL paradigms}
Split learning (SL) moves part of the model to the server to reduce client burden, but can be sensitive to client ordering and harder to scale with many participants. Hybrid designs combine SL partitioning with FL-style parallelism and aggregation. SplitFed formalizes this combination and reports improved scalability and robustness under non-IID data \cite{_5}. Edge-focused analyses argue hybridization can reduce bottlenecks when pure FL or pure SL is used in isolation, especially under bandwidth and compute asymmetry \cite{_6}. More recent work adds architectural controls such as private normalization and decomposition to improve performance under heterogeneity while reducing leakage from intermediate representations \cite{_7}.

Many hybrid FL--SL studies remain benchmark-centric, while clinical decision support often requires calibration, subgroup checks, and robustness under distribution shift. Treatment optimization further shifts the target from prediction to incremental benefit, which increases sensitivity to confounding and selection effects in observational data.

\subsection{Privacy and security risks in collaborative learning}
Distributed training is not inherently private. Information may leak through gradients, model updates, or intermediate activations, and adversaries can exploit auxiliary knowledge or adaptive strategies. A healthcare-focused review summarizes threat models for FL and discusses defenses such as secure aggregation, differential privacy, and cryptographic mechanisms, along with their overhead and practical limitations \cite{_4}. Membership inference audits whether shared artifacts reveal training participation, and recent work shows stronger attacks in federated settings, including multi-round gradient-based strategies \cite{_11,_12}. In SL, risks can be more direct because intermediate activations may enable reconstruction; recent attacks demonstrate recovery of private features even when raw data are never shared \cite{_9,_10}. Surveys of privacy-preserving split learning further describe how cut-layer placement and activation protection trade off with accuracy and overhead \cite{_8}. For clinical collaboration, partitioning is therefore a security-relevant hyperparameter and evaluation should include at least one concrete leakage probe.

\subsection{From prediction to treatment effect and uplift}
Risk prediction does not quantify the incremental benefit of treatment for a specific patient. Heterogeneous treatment effect estimation and uplift modeling target individualized or subgroup-specific benefit. A clinical pharmacology review outlines identification and generalization challenges for CATE estimation and emphasizes sensitivity to dataset shift \cite{_15}. In health-technology assessment, scoping work notes persistent issues with time-varying confounding, uncertainty quantification, and decision-facing evaluation \cite{_16}. Benchmarking studies show that rankings depend on data-generating conditions and evaluation choices, motivating consistent reporting using uplift curves and AUUC-like summaries \cite{_17}. Deep learning studies further suggest clinically meaningful heterogeneity where population averages can hide actionable subgroups \cite{_18}. Broad reviews stress that moving from prediction to prescription requires alignment among objectives, assumptions, and deployment constraints \cite{_19}. Together, these observations motivate uplift evaluation in collaborative learning, where distribution shift and governance constraints can undermine decision reliability across sites.

\subsection{Taxonomy of prior work and the remaining gap}
Table~\ref{tab:taxonomy} summarizes post-2020 work by what institutions exchange and what the model optimizes. Prediction-centric collaboration is mature, while treatment-effect-focused collaboration is less explored, particularly for hybrid FL--SL with explicit privacy auditing and cross-dataset validation.

\begin{table*}[t]
\vspace*{0.12in}
\centering
\small
\setlength{\tabcolsep}{5pt}
\renewcommand{\arraystretch}{1.05}
\caption{Taxonomy of recent collaborative learning in healthcare and the gap addressed in this work.}
\label{tab:taxonomy}
\begin{tabularx}{\textwidth}{>{\raggedright\arraybackslash}p{2.1cm}>{\raggedright\arraybackslash}p{3.0cm}>{\raggedright\arraybackslash}X>{\raggedright\arraybackslash}X}
\toprule
\textbf{Paradigm} & \textbf{Exchanged Objects} & \textbf{Post-2020 Examples} & \textbf{Decision-Support Gap} \\
\midrule
Federated Learning (FL) & Model updates; optional secure aggregation & Challenges \cite{_1}; healthcare evidence \cite{_2}; EHR review \cite{_3} & Often predictive-only; leakage auditing is inconsistent in applied pipelines \cite{_4}. \\
\addlinespace
Split Learning (SL) & Cut-layer activations and backpropagated gradients & SL survey \cite{_8}; representation attacks \cite{_9,_10} & Privacy/compute focus; decision utility such as uplift prioritization is rarely evaluated. \\
\addlinespace
Hybrid FL and SL & Split representations with federated aggregation & SplitFed \cite{_5}; edge analysis \cite{_6}; PPSFL \cite{_7} & Limited treatment-optimization focus; unified auditing and cross-dataset validation are often incomplete. \\
\addlinespace
Causal / Uplift (Centralized) & Centralized $(X,T,Y)$ with counterfactual estimators & CATE challenges \cite{_15}; HTA review \cite{_16}; uplift benchmarking \cite{_17} & Typically assumes pooled data; governance and multi-site constraints are not addressed. \\
\midrule
\textbf{This Work} & \textbf{Hybrid FL and SL with uplift evaluation} & \textbf{Proposed framework} & \textbf{Unified utility, communication, and membership inference auditing across datasets and client partitions.} \\
\bottomrule
\end{tabularx}
\end{table*}

Overall, prior work provides strong components for collaboration and treatment-effect estimation, but their integration remains limited. Hybrid FL--SL systems are rarely evaluated on incremental-effect decision targets and often omit leakage probing aligned with modern attacks \cite{_9,_11}. Conversely, uplift methods are well studied in centralized settings, yet their assumptions are seldom revisited under cross-site heterogeneity and collaborative constraints \cite{_16,_19}. Our work targets this gap by evaluating a hybrid FL--SL pipeline for uplift-based prioritization and reporting a unified utility, communication, and privacy profile across datasets and client partitions.


\section{Methodology}
\label{sec:methodology}

\subsection{Problem setting and decision objective}
We study privacy-preserving collaborative learning for clinical decision support when institutions cannot pool patient-level data. Client $k\in\{1,\dots,K\}$ holds $\mathcal{D}_k=\{(x_i,t_i,y_i)\}_{i=1}^{n_k}$, where $x\in\mathbb{R}^d$ are covariates, $t\in\{0,1\}$ is a treatment or exposure proxy, and $y\in\{0,1\}$ is a binary outcome. Training exchanges model updates (federated learning), intermediate activations (split learning), or both (hybrid).

Beyond factual risk estimation, we prioritize limited interventions toward individuals expected to benefit. With potential outcomes $(Y(1),Y(0))$, the conditional average treatment effect (CATE) is
\begin{equation}
\tau(x)=\mathbb{E}[Y(1)-Y(0)\mid X=x].
\label{eq:cate}
\end{equation}
Because $\tau(x)$ is unobserved and $T$ is observational, we estimate $\mu_1(x)=\mathbb{P}(Y=1\mid X=x,T=1)$ and $\mu_0(x)=\mathbb{P}(Y=1\mid X=x,T=0)$ and define uplift as
\begin{equation}
\widehat{\tau}(x)=\widehat{\mu}_1(x)-\widehat{\mu}_0(x).
\label{eq:uplift}
\end{equation}
We use $\widehat{\tau}(x)$ for top-$q$ ranking rather than causal claims; with proxy treatments, AUUC may be negative because it integrates uplift over coverage. We therefore include a random-ranking baseline and report top-$q$ uplift for capacity-limited actionability.

\subsection{Datasets, client partitions, and preprocessing}
We evaluate on three public clinical datasets. Each dataset is harmonized into a unified schema $(X,T,Y)$ and a non-IID client identifier \texttt{client\_id} that mimics institution-like heterogeneity. The same modeling code and evaluation protocol are reused across datasets to support cross-dataset validation under distribution shift. Table~\ref{tab:dataset_schema} summarizes how each dataset is mapped into $(X,T,Y)$ and how \texttt{client\_id} is defined to induce non-IID collaboration.

\begin{table*}[t]
\vspace*{0.12in}
\centering
\small
\setlength{\tabcolsep}{5pt}
\renewcommand{\arraystretch}{1.2}
\caption{Unified uplift construction across datasets. $T$ denotes a treatment or exposure proxy; $Y$ is the outcome; \texttt{client\_id} defines a non-IID collaborative split.}
\label{tab:dataset_schema}
\begin{tabularx}{\textwidth}{>{\raggedright\arraybackslash}p{2.3cm} >{\raggedright\arraybackslash}p{3.6cm} >{\raggedright\arraybackslash}p{3.4cm} >{\raggedright\arraybackslash}X}
\hline
\textbf{Dataset} & \textbf{$T$ (treatment proxy)} & \textbf{$Y$ (outcome)} & \textbf{Client partition (\texttt{client\_id})} \\
\hline
eICU-demo \cite{_21} & Early vasopressor exposure in first 6 hours (binary) & In-hospital mortality (binary) & Admission diagnosis groups (top diagnoses vs other), inducing site-like heterogeneity \\
MEPS 2022 \cite{_22} & Statin exposure from prescribed medicine records (binary) & Inpatient utilization indicator (binary) & Region-based partition to mimic policy and access differences \\
NHANES 2015--2018 \cite{_23} & Statin mention in prescription file (binary) & Controlled cholesterol indicator (binary) & Survey cycle-based partition (cycle I vs J) for temporal shift \\
\hline
\end{tabularx}
\end{table*}

All datasets are converted to numeric feature matrices. We apply median imputation for missing numeric values (fit on the training split only) and standardize features using training-split statistics to prevent leakage. Splits are stratified on $Y$ into train, validation, and test partitions and are reused across all collaborative modes to ensure fair comparison.

\subsection{Uplift modeling}
\subsubsection{Two-head uplift network}
We adopt a two-head architecture with a shared trunk $f_{\theta}:\mathbb{R}^d\rightarrow\mathbb{R}^h$ and treatment-specific heads $g_{\phi_1},g_{\phi_0}$. For an input $x$,
\begin{equation}
z=f_{\theta}(x),\qquad
\widehat{\mu}_1(x)=\sigma(g_{\phi_1}(z)),\qquad
\widehat{\mu}_0(x)=\sigma(g_{\phi_0}(z)),
\label{eq:twohead}
\end{equation}
where $\sigma(\cdot)$ is the logistic sigmoid. The factual probability used for supervised training is
\begin{equation}
\widehat{p}(x,t)=t\widehat{\mu}_1(x)+(1-t)\widehat{\mu}_0(x).
\label{eq:factual_prob}
\end{equation}
Parameters are optimized by minimizing the empirical binary cross-entropy on observed outcomes,
\begin{align}
\min_{\theta,\phi_0,\phi_1}\; &\sum_{(x_i,t_i,y_i)} \ell\!\left(y_i,\widehat{p}(x_i,t_i)\right), \label{eq:factual_loss_min} \\
\ell(y,p) &= -y\log p-(1-y)\log(1-p). \label{eq:factual_loss_def}
\end{align}
This objective is compatible with centralized, federated, split, and hybrid training because it requires only observed $(x,t,y)$.

\subsubsection{Doubly robust sanity-check baseline}
As a diagnostic reference, we also compute a doubly robust (DR) pseudo-effect using a propensity model $\widehat{e}(x)=\mathbb{P}(T=1\mid X=x)$ and outcome models $(\widehat{\mu}_1,\widehat{\mu}_0)$. For sample $(x_i,t_i,y_i)$,
\begin{equation}
\begin{aligned}
\hat{\tau}_{DR}(x_i)
&= \left(\hat{\mu}_1(x_i)-\hat{\mu}_0(x_i)\right) \\
&\quad + \frac{t_i-\hat{e}(x_i)}{\hat{e}(x_i)\left(1-\hat{e}(x_i)\right)}
\left(y_i-\hat{\mu}_{t_i}(x_i)\right).
\end{aligned}
\label{eq:dr}
\end{equation}

where $\widehat{\mu}_{t_i}(x_i)=t_i\widehat{\mu}_1(x_i)+(1-t_i)\widehat{\mu}_0(x_i)$. We use this estimator for debugging and ranking-signal validation, not as a formal causal claim.

\subsection{Collaborative training protocols}
We compare four collaboration modes: Centralized, Federated Learning (FL), Split Learning (SL), and Hybrid FL-SL. These settings represent common deployment choices and hybrid designs discussed in recent work \cite{_1,_5,_6}. Table~\ref{tab:modes} outlines the four training modes and the main object exchanged between clients and the coordinating server.

\begin{table*}[t]
\centering
\small
\setlength{\tabcolsep}{5pt}
\renewcommand{\arraystretch}{1.2}
\caption{Compared training modes. ``Exchange'' indicates the primary object communicated across clients and server.}
\label{tab:modes}
\begin{tabularx}{\textwidth}{>{\raggedright\arraybackslash}p{2.5cm} >{\raggedright\arraybackslash}p{3.6cm} >{\raggedright\arraybackslash}p{3.9cm} >{\raggedright\arraybackslash}X}
\hline
\textbf{Mode} & \textbf{Exchange} & \textbf{Computation placement} & \textbf{Key rationale} \\
\hline
Centralized & None (pooled data) & Single trainer & Upper-bound utility reference \\
FL (FedAvg) & Model weights/updates & Full model trained locally & Keeps raw data local; aggregates updates \cite{_1} \\
SL & Activations and gradients at cut layer & Trunk on client, head on server & Reduces client compute; shifts depth to server \\
Hybrid FL-SL & Split activations plus federated aggregation of client trunks & Trunk local (federated), head server (split) & Combines parallelism and partitioning \cite{_5,_6} \\
\hline
\end{tabularx}
\end{table*}

\subsubsection{Federated learning with FedAvg}
In FL, each client performs local training for $E$ epochs starting from global parameters $w^{(r)}$ at round $r$. The server aggregates client parameters via weighted averaging,
\begin{equation}
w^{(r+1)}=\sum_{k=1}^{K}\frac{n_k}{\sum_{j=1}^{K}n_j}\,w_k^{(r,E)}.
\label{eq:fedavg}
\end{equation}
FedAvg is used as the primary FL baseline because it is standard and interpretable under heterogeneity \cite{_1}.

\subsubsection{Split learning}
In SL, the model is split at a cut layer: the client computes $z=f_{\theta}(x)$ and transmits $z$ to the server, which applies the heads, computes the loss in \eqref{eq:factual_loss_def}, and returns gradients with respect to $z$. The client completes backpropagation through $f_{\theta}$. Since the server observes representations, the cut location is treated as a privacy-relevant design choice \cite{_8,_9}.

\subsubsection{Hybrid FL-SL with optional client personalization}
Hybrid training combines split execution with federated aggregation of the client-side trunk. Each client maintains a trunk $\theta_k$ and may include a lightweight local adapter $a_k$ that is not shared. The adapter applies a residual correction at the cut representation,
\begin{equation}
z=f_{\theta}(x),\qquad
\tilde{z}=z+a_k(z).
\label{eq:adapter}
\end{equation}
Adapters support client-specific adjustment under non-IID partitions while the shared trunk captures global structure. Algorithm~\ref{alg:hybrid} summarizes the full hybrid FL-SL procedure, including optional client adapters and the FedAvg aggregation of shared trunk parameters.

\begin{algorithm}[t]
\vspace{8pt}
\caption{Hybrid FL-SL training with optional client adapters}
\label{alg:hybrid}
\footnotesize
\begin{algorithmic}[1]
\Require Clients $\{\mathcal{D}_k\}_{k=1}^{K}$, rounds $R$, local steps $E$, learning rates $\eta_c,\eta_s$
\Ensure Shared trunk $\theta^{(R)}$, server head $\phi^{(R)}$, and local adapters $\{a_k\}$ (if used)
\State Initialize shared trunk base $\theta^{(0)}$ and server head $\phi^{(0)}$
\For{$r \gets 0$ to $R-1$}
  \State Server broadcasts $\theta^{(r)}$ to all clients
  \ForAll{clients $k$ \textbf{in parallel}}
    \State Set local trunk $\theta_k \gets \theta^{(r)}$ and keep local adapter $a_k$ (if used)
    \For{$e \gets 1$ to $E$}
      \State Client computes $z \gets f_{\theta_k}(x)$ (or $\tilde{z} \gets z + a_k(z)$) and sends to server
      \State Server computes loss $\ell$ (Eq.~(6)) and returns $\nabla_{z}\ell$; server updates head $\phi$ using $\eta_s$
      \State Client updates $\theta_k$ (and $a_k$ if used) using received gradients and $\eta_c$
    \EndFor
    \State Client returns updated shared trunk parameters $\theta_k$ (excluding adapter parameters)
  \EndFor
  \State Server aggregates shared trunk base using FedAvg to obtain $\theta^{(r+1)}$
\EndFor
\State \Return $\theta^{(R)}$, $\phi^{(R)}$, and $\{a_k\}$ if used
\end{algorithmic}
\end{algorithm}

\subsection{Threat model, privacy audit, and mitigation}
Collaborative learning may leak information through exchanged updates or representations even when raw records remain local \cite{_4}. We adopt an honest-but-curious server threat model and focus on empirical auditing rather than certified guarantees.

\subsubsection{Membership inference on cut-layer representations}
For split and hybrid settings, we audit membership inference on cut-layer representations. For a target client $k$, we sample member points from its training split and non-member points from its test split, compute representations $z$, and train a lightweight attacker $A(\cdot)$ to predict membership:
\begin{equation}
\widehat{m}=A(z),\qquad m\in\{0,1\}.
\label{eq:mia}
\end{equation}
Attack performance is summarized by AUC, providing a comparable leakage signal across datasets and training modes \cite{_11,_12}. Algorithm~\ref{alg:mia_audit} formalizes the membership inference audit used to quantify leakage from cut-layer representations in split and hybrid settings.

\subsubsection{Mitigation via clipping and noise}
To explore privacy-utility trade-offs, transmitted activations can be clipped and perturbed:
\begin{equation}
z \leftarrow z\cdot \min\left(1, \frac{c}{\lVert z\rVert_2}\right),\qquad
z \leftarrow z + \sigma \varepsilon,\;\;\varepsilon\sim\mathcal{N}(0,I),
\label{eq:clip_noise}
\end{equation}
where $c$ is the clipping norm and $\sigma$ controls noise magnitude. This is evaluated as an engineering control rather than a formal differential privacy claim \cite{_4,_8}. 

\begin{algorithm}
\caption{Membership inference audit on cut-layer representations}
\label{alg:mia_audit}
\footnotesize
\begin{algorithmic}[1]
\Require Client $k$, trunk model $f_{\theta_k}$, member set $\mathcal{M}$ (train), non-member set $\mathcal{N}$ (test), sample size $m$
\Ensure Attack AUC as the audited leakage signal
\State Sample $m$ records from $\mathcal{M}$ and $m$ records from $\mathcal{N}$
\State Compute $Z_M \gets \{ f_{\theta_k}(x) : x \in \mathcal{M} \}$ and $Z_N \gets \{ f_{\theta_k}(x) : x \in \mathcal{N} \}$
\State Form attack dataset $Z \gets Z_M \cup Z_N$ with labels $s{=}1$ for $Z_M$ and $s{=}0$ for $Z_N$
\State Train attacker $A$ (logistic regression) on $(Z,s)$
\State Evaluate attacker performance by AUC on a held-out split
\State \Return attack AUC
\end{algorithmic}
\end{algorithm}

\subsection{Evaluation protocol and reporting}
We report both predictive utility and decision utility. Predictive utility is summarized by test AUROC using the factual probability in \eqref{eq:factual_prob}. Decision utility is summarized by an AUUC-style integral of the uplift curve computed from $\widehat{\tau}(x)$ in \eqref{eq:uplift}. To reduce sensitivity to extreme propensities in observational data, we estimate $\widehat{e}(x)$ on the training split and trim test points that violate positivity:
\begin{equation}
\mathcal{I}_{\mathrm{keep}}=\left\{i:\alpha \le \widehat{e}(x_i)\le 1-\alpha\right\}.
\label{eq:trim}
\end{equation}
We report the retained fraction as a diagnostic and evaluate uplift on $\mathcal{I}_{\mathrm{keep}}$. For split and hybrid modes, we also report membership inference AUC as the privacy audit signal. Communication cost is estimated in bytes per round: FL scales with parameter exchange, while SL and Hybrid additionally include cut-layer activations and gradients \cite{_6}. All methods reuse identical splits and preprocessing. Hyperparameters are selected on validation data and then fixed for test evaluation. Results are aggregated across multiple random seeds (mean and standard deviation).

\subsection{Research framework overview}
\begin{figure*}[t]
\centering
\includegraphics[width=0.97\linewidth]{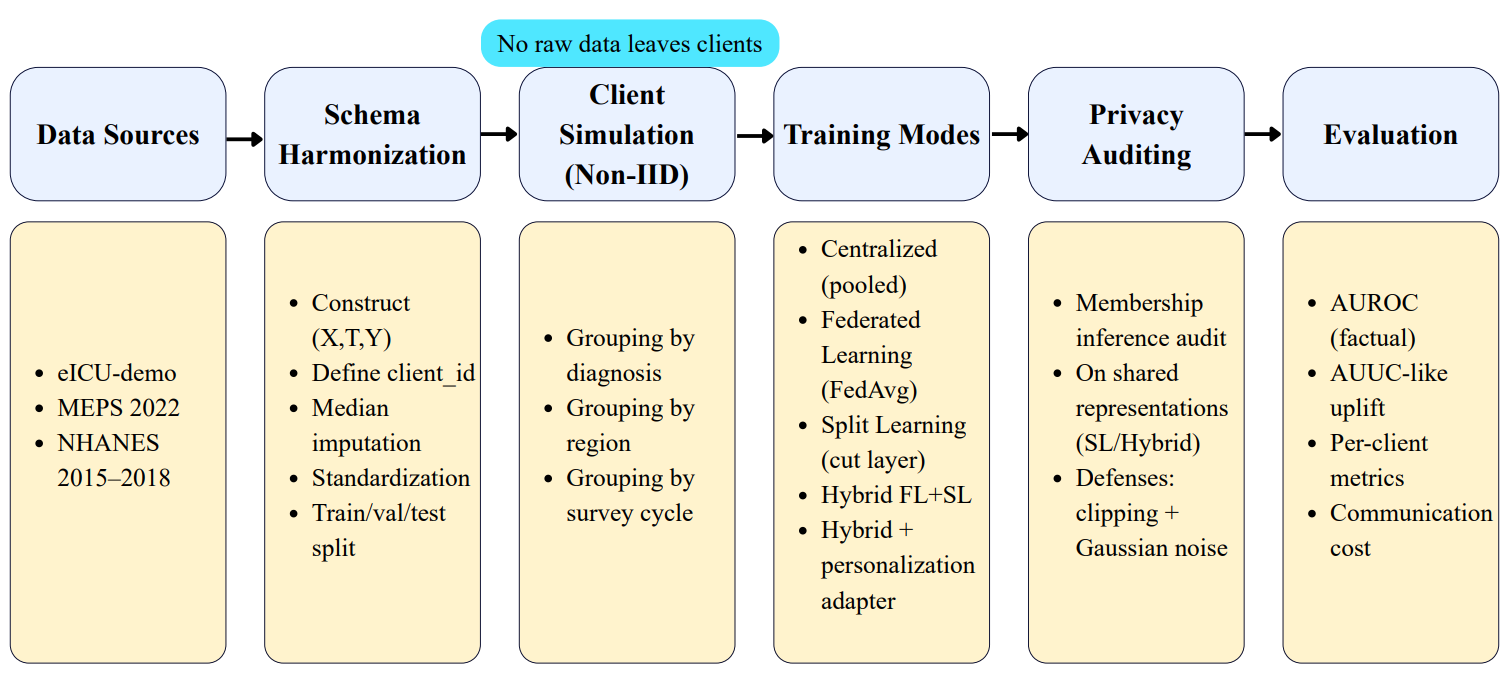}
\caption{End-to-end research framework for privacy-preserving collaborative uplift modeling.}
\label{fig:framework}
\end{figure*}

Figure~\ref{fig:framework} summarizes the end-to-end pipeline: harmonization into $(X,T,Y)$, client partitioning, collaborative training across modes, privacy auditing, and unified evaluation.

\subsection{Implementation notes}
All pipelines are implemented in Python. The two-head uplift network is trained with Adam using mini-batches; propensity estimation and the membership attacker use logistic regression. Table~\ref{tab:repro_settings} lists the fixed architecture and hyperparameters used to reproduce our experiments. Communication rounds, local steps, and defense settings follow the configurations reported in Table~\ref{tab:table2_main}. Experiments run on GPU when available for neural optimization; preprocessing and evaluation run efficiently on CPU.

\begin{table}[t]
\vspace*{0.12in}
\centering
\caption{Reproducibility settings used in all experiments. Training rounds $R$ follow Table~\ref{tab:table2_main}.}
\label{tab:repro_settings}
\small
\setlength{\tabcolsep}{4pt}
\renewcommand{\arraystretch}{1.08}
\begin{tabular}{p{3.2cm} p{4.7cm}}
\hline
Client trunk & 2-layer MLP: $d \rightarrow 64 \rightarrow 32$ (ReLU) \\
Cut layer & After trunk output $z \in \mathbb{R}^{32}$ (Split/Hybrid) \\
Server heads & Two independent heads: $32 \rightarrow 1$ + sigmoid \\
Training & Adam; $\eta_c=\eta_s=10^{-3}$; batch $=256$ (train), $512$ (eval); local $E=1$ \\
Defense & Activation clipping $c=1.0$; Gaussian noise $\sigma=0.05$ \\
\hline
\end{tabular}
\end{table}

\section{Results and Discussion}
\label{sec:results}

We evaluate four deployment-relevant axes: predictive utility (factual discrimination), decision utility under capacity constraints (uplift ranking), privacy leakage at the collaboration boundary (membership inference on shared representations), and deployment cost (communication). Experiments follow Section~\ref{sec:methodology}, including positivity trimming and membership auditing for split and hybrid modes. Because treatments are observational proxies, uplift is interpreted as a decision-ranking signal under a fixed protocol rather than a causal effect.

\subsection{Main performance across datasets}
\label{subsec:main_summary}

Table~\ref{tab:table2_main} reports test-set results for eICU, MEPS, and NHANES under Centralized, FedAvg, Split, Hybrid, Hybrid+Personalization, and Hybrid+Defense. We report AUROC (factual prediction), AUUC (uplift ranking), and worst-client AUUC (non-IID robustness). We also include trimming rate and end-of-curve uplift as diagnostics. End-of-curve uplift reflects the treated minus control gap at full coverage and is largely method-invariant; with observational proxies it may be negative, so AUUC can also be negative without implying failure. To verify actionability, we include a random-ranking baseline AUUC (Table~\ref{tab:table2_main}); learned uplift rankings should exceed random even when the global end uplift is negative. We additionally compare against a \emph{random uplift ranking} baseline (by permuting predicted uplift scores) to verify that learned rankings provide non-trivial prioritization beyond chance.

\begin{table*}[t]
\centering
\caption{Main results across datasets. AUROC: factual discrimination. AUUC: uplift ranking under trimming (may be negative with observational proxies). Worst-client AUUC: non-IID robustness. Comm and rounds: deployment cost; MIA AUC: leakage audit for split/hybrid modes.}

\label{tab:table2_main}
\footnotesize
\setlength{\tabcolsep}{4pt}
\begin{tabularx}{\textwidth}{|>{\raggedright\arraybackslash}X|>{\raggedright\arraybackslash}X|S[table-format=1.2]|S[table-format=1.2]|S[table-format=1.2]|S[table-format=2.2]|S[table-format=1.2]|S[table-format=2.2]|S[table-format=2.0]|>{\raggedright\arraybackslash}X|}
\hline
\textbf{Dataset} & \textbf{Method} & \textbf{AUROC} & \textbf{AUUC} & \textbf{End} & \textbf{Trim} & \textbf{Worst} & \textbf{Comm} & \textbf{Rounds} & \textbf{MIA} \\
 & & & & \textbf{Uplift} & \% & \textbf{AUUC} & \textbf{(MB)} & & \textbf{AUC} \\
\hline
eicu & Centralized & 0.51 & -0.03 & -0.29 & 9.92 & -0.81 & 0.00 & 5 & N/A \\
eicu & FedAvg & 0.48 & -0.11 & -0.29 & 9.92 & -0.37 & 2.04 & 8 & N/A \\
eicu & Split & 0.51 & -0.13 & -0.29 & 9.92 & -0.72 & 3.61 & 4 & 0.49 \\
eicu & Hybrid & 0.49 & -0.10 & -0.29 & 9.92 & -0.06 & 3.81 & 4 & 0.48 \\
eicu & Hybrid+Pers. & 0.54 & -0.29 & -0.29 & 9.92 & -0.71 & 3.81 & 4 & 0.50 \\
eicu & Hybrid+Def. & 0.48 & -0.21 & -0.29 & 17.06 & -0.38 & 3.81 & 4 & 0.42 \\
\hline
meps & Centralized & 0.64 & -0.07 & -0.09 & 10.01 & -0.10 & 0.00 & 5 & N/A \\
meps & FedAvg & 0.66 & -0.07 & -0.09 & 10.01 & -0.11 & 0.85 & 4 & N/A \\
meps & Split & 0.62 & -0.11 & -0.09 & 10.01 & -0.14 & 32.16 & 4 & 0.51 \\
meps & Hybrid & 0.75 & -0.09 & -0.09 & 10.01 & -0.11 & 80.80 & 10 & 0.49 \\
meps & Hybrid+Pers. & 0.73 & -0.10 & -0.09 & 10.01 & -0.14 & 56.56 & 7 & 0.52 \\
meps & Hybrid+Def. & 0.67 & -0.12 & -0.09 & 10.01 & -0.15 & 32.32 & 4 & 0.52 \\
\hline
nhanes & Centralized & 0.75 & 0.08 & 0.02 & 9.99 & 0.08 & 0.00 & 9 & N/A \\
nhanes & FedAvg & 0.77 & 0.09 & 0.02 & 9.99 & 0.08 & 0.78 & 10 & N/A \\
nhanes & Split & 0.75 & 0.08 & 0.02 & 9.99 & 0.08 & 33.23 & 6 & 0.51 \\
nhanes & Hybrid & 0.76 & 0.08 & 0.02 & 9.99 & 0.07 & 33.29 & 6 & 0.48 \\
nhanes & Hybrid+Pers. & 0.78 & 0.09 & 0.02 & 9.99 & 0.08 & 55.48 & 10 & 0.50 \\
nhanes & Hybrid+Def. & 0.76 & 0.08 & 0.02 & 9.99 & 0.07 & 55.48 & 10 & 0.50 \\
\hline
\end{tabularx}
\end{table*}

NHANES shows consistently positive uplift signals and stable ranking behavior, whereas eICU and MEPS have negative dataset-level treated minus control gaps under the chosen observational proxies. In these proxy settings, the practical objective is whether the model can prioritize comparatively better candidates under limited capacity, not whether the global average gap is positive.

\subsection{Decision utility under capacity constraints}
\label{subsec:decision_utility}

\begin{figure*}[t]
\centering
\includegraphics[width=0.32\linewidth]{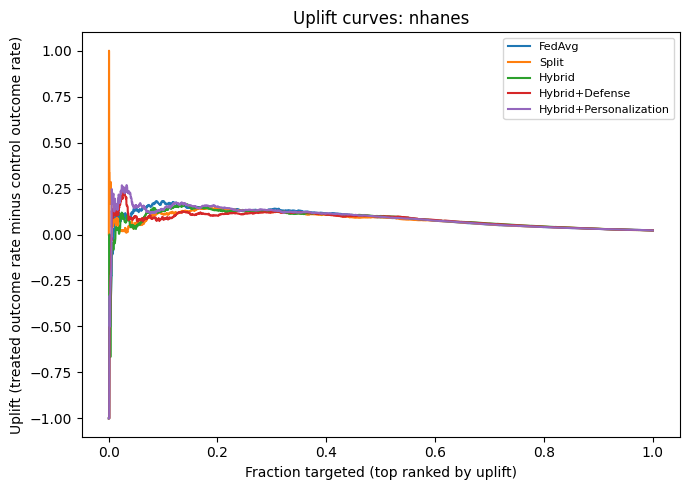}
\includegraphics[width=0.32\linewidth]{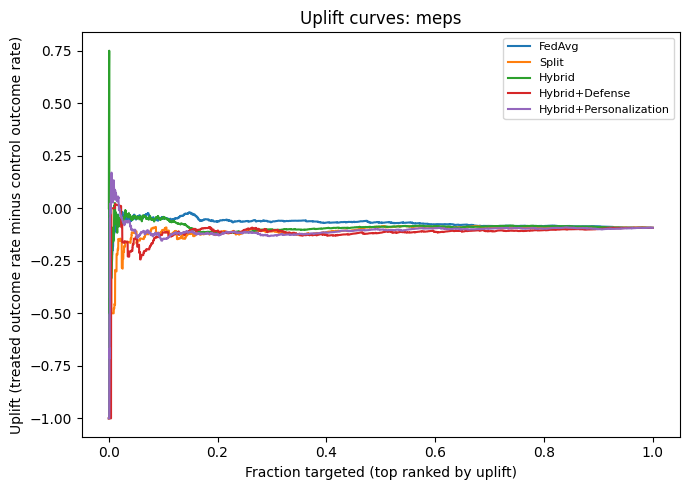}
\includegraphics[width=0.32\linewidth]{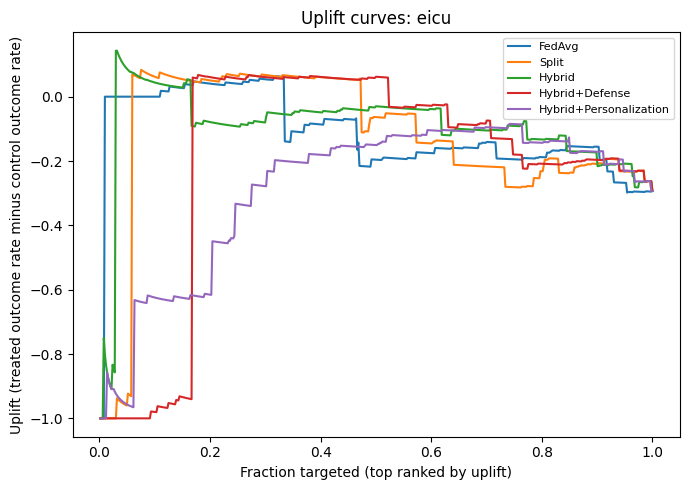}
\caption{Uplift curves under capacity targeting (top fraction by predicted uplift).}
\label{fig:uplift_curves}
\end{figure*}

Figure~\ref{fig:uplift_curves} reports uplift curves, where patients are ranked by predicted uplift and the curve tracks treated minus control outcome rate within the targeted prefix. NHANES exhibits a clear positive early-targeting region and a smooth decay toward the full-coverage gap. For eICU and MEPS, curves trend negative at larger targeting fractions, so AUUC can be negative even when early targeting provides a useful ranking signal; interpretation should therefore emphasize the top-fraction behavior together with AUUC.

\subsection{Privacy leakage and privacy--utility trade-off}
\label{subsec:privacy_results}

\begin{figure}[t]
\centering
\includegraphics[width=0.98\linewidth]{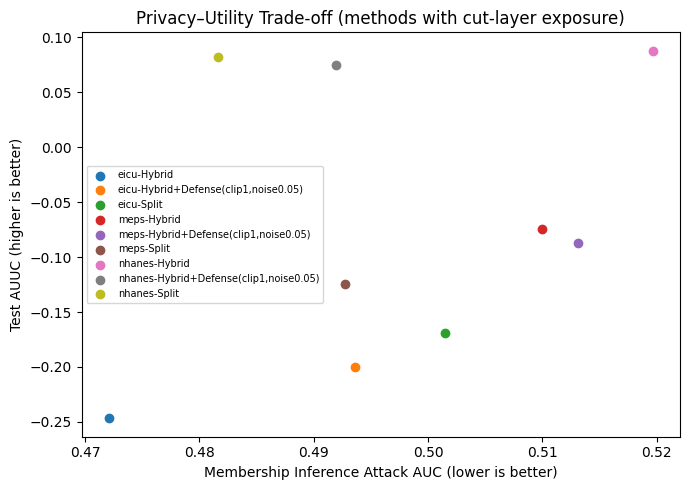}
\caption{Privacy--utility trade-off: AUUC vs membership inference AUC for defended configurations.}
\label{fig:privacy_utility}
\end{figure}

Privacy leakage is audited using membership inference AUC on cut-layer representations for Split and Hybrid variants (Table~\ref{tab:table2_main}). Values near $0.5$ indicate that the tested lightweight attacker performs at chance in these settings, but this is an empirical audit signal rather than a proof of privacy (stronger attackers, different cut layers, or larger models can change leakage). We therefore report defended operating points (clipping/noise) to show how design choices shift the privacy--utility trade-off in Figure~\ref{fig:privacy_utility}.

\subsection{Deployment cost and non-IID robustness}
\label{subsec:efficiency_robustness}

Deployment cost is captured by total communication and training rounds in Table~\ref{tab:table2_main}. FedAvg is lighter because it exchanges parameters once per round ($\mathcal{O}(|w|)$), while split/hybrid modes additionally transmit cut-layer activations and gradients (roughly $\mathcal{O}(B\cdot d_{\text{cut}})$), which can dominate payload. From an operational perspective, the hybrid design is most relevant when central pooling is infeasible (policy/contract) and when reduced governance overhead and boundary auditing justify higher bandwidth; in practice, stragglers and scale can be handled via partial participation, fewer rounds, and activation compression.

Robustness under non-IID client partitions is summarized in Table~\ref{tab:table6_fair}. We report mean and dispersion across clients together with worst-client AUUC and AUROC, since averaged scores can hide failure modes under heterogeneous client distributions.

\begin{table*}[t]
\centering
\caption{Non-IID robustness across clients. Mean and standard deviation are computed across clients; worst-client metrics highlight conservative deployment risk.}
\label{tab:table6_fair}
\footnotesize
\setlength{\tabcolsep}{3pt}
\begin{tabularx}{\textwidth}{|>{\raggedright\arraybackslash}X|>{\raggedright\arraybackslash}X|S[table-format=-1.4]|S[table-format=1.4]|S[table-format=-1.4]|S[table-format=1.4]|S[table-format=1.4]|S[table-format=1.4]|}
\hline
\multicolumn{2}{|c|}{\textbf{Configuration}} & \multicolumn{3}{c|}{\textbf{AUUC across clients}} & \multicolumn{3}{c|}{\textbf{AUROC across clients}} \\
\cline{3-8}
\textbf{Dataset} & \textbf{Method} & \textbf{Mean} & \textbf{Std} & \textbf{Worst} & \textbf{Mean} & \textbf{Std} & \textbf{Worst} \\
\hline
eicu & Centralized & -0.4650 & 0.3288 & -0.7938 & 0.3730 & 0.1674 & 0.2056 \\
eicu & FedAvg & -0.2521 & 0.2334 & -0.4855 & 0.3045 & 0.1045 & 0.2000 \\
eicu & Split & -0.3475 & 0.0180 & -0.3655 & 0.3167 & 0.2056 & 0.1111 \\
eicu & Hybrid & -0.4282 & 0.0871 & -0.5153 & 0.4844 & 0.1177 & 0.3667 \\
eicu & Hybrid+Pers. & -0.2799 & 0.1645 & -0.4444 & 0.3276 & 0.2443 & 0.0833 \\
eicu & Hybrid+Def. & -0.2852 & 0.1450 & -0.4301 & 0.3118 & 0.2341 & 0.0778 \\
\hline
meps & Centralized & -0.0867 & 0.0344 & -0.1431 & 0.6721 & 0.0434 & 0.6033 \\
meps & FedAvg & -0.0698 & 0.0256 & -0.0976 & 0.6947 & 0.0323 & 0.6461 \\
meps & Split & -0.0901 & 0.0199 & -0.1158 & 0.7181 & 0.0268 & 0.6926 \\
meps & Hybrid & -0.0711 & 0.0299 & -0.1075 & 0.7310 & 0.0306 & 0.6996 \\
meps & Hybrid+Pers. & -0.0885 & 0.0168 & -0.1143 & 0.7197 & 0.0284 & 0.6899 \\
meps & Hybrid+Def. & -0.1087 & 0.0303 & -0.1434 & 0.6749 & 0.0432 & 0.6246 \\
\hline
nhanes & Centralized & 0.0923 & 0.0081 & 0.0842 & 0.7673 & 0.0047 & 0.7626 \\
nhanes & FedAvg & 0.0793 & 0.0002 & 0.0791 & 0.7331 & 0.0004 & 0.7327 \\
nhanes & Split & 0.0734 & 0.0004 & 0.0730 & 0.7292 & 0.0021 & 0.7270 \\
nhanes & Hybrid & 0.0882 & 0.0023 & 0.0859 & 0.7699 & 0.0038 & 0.7661 \\
nhanes & Hybrid+Pers. & 0.0862 & 0.0041 & 0.0822 & 0.7727 & 0.0053 & 0.7673 \\
nhanes & Hybrid+Def. & 0.0771 & 0.0095 & 0.0676 & 0.7375 & 0.0082 & 0.7294 \\
\hline
\end{tabularx}
\end{table*}

\subsection{Discussion}
\label{sec:discussion}

This paper frames hybrid federated plus split learning as a practical systems and privacy approach for healthcare decision support under governance constraints and heterogeneous client populations. Across datasets and client partitions, our results support four deployment-relevant takeaways \cite{_1,_4}.

First, hybridization is most useful when the objective is decision-facing and clients are heterogeneous. FedAvg is communication-efficient but can yield unstable uplift ranking under client shifts \cite{_1}. Split learning centralizes the head and can reduce client-side burden, but intermediate representations expand the privacy surface \cite{_9,_10}. Hybrid FL--SL aggregates client trunks while retaining a split boundary, providing a clear interface where privacy controls can be applied and audited \cite{_5,_6}.

Second, uplift metrics must be interpreted under observational proxy treatments. In eICU and MEPS, the treated minus control gap is negative under the chosen proxies, so AUUC can be negative because it integrates the full targeting range. The appropriate claim is consistent prioritization under a fixed protocol and capacity constraint, not recovery of positive causal effects \cite{_15,_17}.

Third, privacy is best treated as a trade-off rather than a guarantee. Membership inference AUC provides an audit signal, and activation clipping with noise can reduce audited leakage with a measurable utility impact \cite{_4,_11}. This motivates reporting privacy and utility jointly and selecting operating points based on deployment requirements.

Fourth, deployability depends on cost and robustness, not only on averages. Communication can increase in split-based methods when activation and gradient exchanges dominate payload \cite{_6}, and worst-client reporting shows that means can hide failures under non-IID partitions \cite{_1}. Overall, hybrid FL--SL provides a practical design space for privacy-preserving collaborative decision support.

\section{Conclusion}
\label{sec:conclusion}

We presented a hybrid FL--SL framework for collaborative healthcare decision support when patient records cannot be centrally pooled. By keeping client-side trunks local and placing the prediction head on the server, the approach enables cross-client learning while making the collaboration boundary explicit. Across three public datasets, hybrid variants deliver competitive prediction quality and practical uplift-based ranking under capacity constraints. We further audit privacy via membership inference on cut-layer representations and show that activation clipping with Gaussian noise can reduce the audited attack signal with modest utility impact, making privacy a tunable design choice. Future work will emphasize site-level validation, stronger adversarial audits, and objectives that better align uplift ranking with causal and calibration requirements in clinical workflows.

\balance

\bibliographystyle{IEEEtran}
\bibliography{Sample-base}

\end{document}